\documentclass[11pt,a4paper]{article}
\usepackage[hyperref]{eacl2021}
\usepackage{times}
\usepackage{latexsym}

\usepackage{graphics}
\usepackage{graphicx}
\usepackage{microtype}
\usepackage{caption, subcaption} 

\aclfinalcopy 

\title{UVCE-IIITT@DravidianLangTech-EACL2021: Tamil Troll Meme Classification: You need to Pay more Attention}

\author{Siddhanth U Hegde\(^1\), Adeep Hande\(^2\), Ruba Priyadharshini\(^3\), \\\textbf{Sajeetha Thavareesan\(^4\)}, \textbf{Bharathi Raja Chakravarthi\(^5\)} \\  \(^1\) University Visvesvaraya College of Engineering, Bangalore University, \\  \(^2\)Indian Institute of Information Technology Tiruchirappalli, Tamil Nadu, \\ \(^3\)ULTRA Arts and Science College, India,\(^4\)Eastern University, Sri Lanka\\  \(^5\)National University of Ireland Galway \\   {\tt siddhanthhegde227@gmail.com}} 
\date{}

\begin{document}
\maketitle
\begin{abstract}
 Tamil is a Dravidian language that is commonly used and spoken in the southern part of Asia. In the era of social media, memes have been a fun moment in the day-to-day life of people. Here, we try to analyze the true meaning of Tamil memes by categorizing them as troll and non-troll. We propose an ingenious model comprising of a transformer-transformer architecture that tries to attain state-of-the-art by using attention as its main component. The dataset consists of troll and non-troll images with their captions as text. The task is a binary classification task. The objective of the model is to pay more attention to the extracted features and to ignore the noise in both images and text.
\end{abstract}
 
\section{Introduction} 
Over the past decade, memes have become a ubiquitous phenomenon over the internet. Memes can come in several formats such as images, video, etc. Memes can take a combined form of both text and images too. Due to its vast popularity, different people perceive memes distinctively. Recent studies have prompted the usage of memes as a mode of communication across social media platforms. The presence of text in images makes it harder to decode the sentiment or any other characteristic \cite{avvaru-vobilisetty-2020-bert}. Regardless of the type of the meme, they may be changed, recreated over social media networks, and tend to be used in contexts involving sensitive topics such as politics, casteism, etc, to add a sarcastic perspective \cite{8354676,Nave2018TalkingIP}.
\begin{figure*}[!h]
\centering
\includegraphics[width=\textwidth,height=8cm]{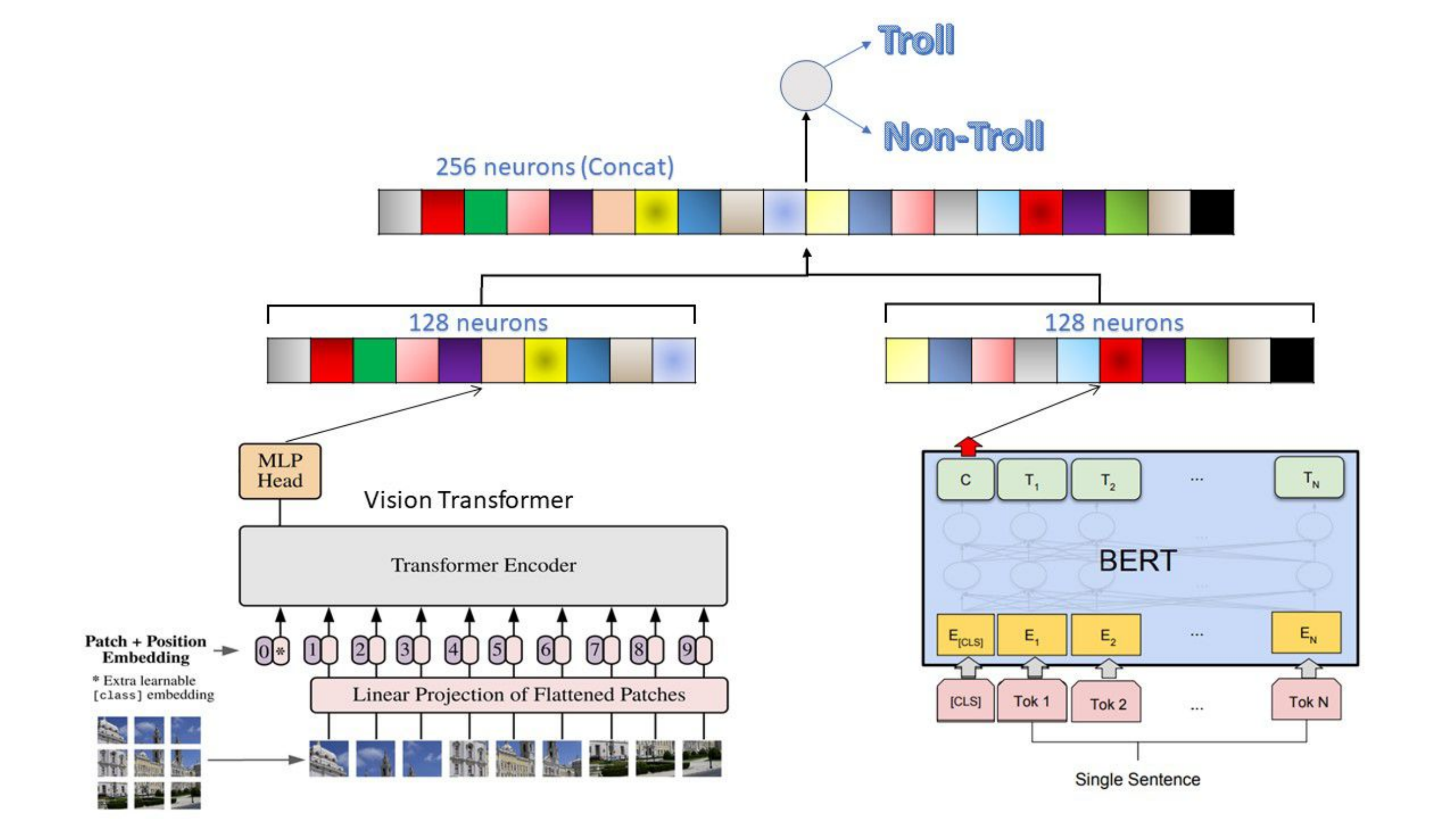}
\caption{System Architecture \cite{dosovitskiy2021an,devlin-etal-2019-bert}} \label{fig1}
\end{figure*}
Due to its multimodality, conscientious analysis of memes can shed light on the societal factors, their implications on culture, and the values promoted by them \cite{Milner2013FCJ156HT}. In addition to that, analyzing the intended emotion of a meme could help us acknowledge fake news, offensive content that is being propagated using the internet memes as a medium, thus helping in eradicating the spread of misinformation and hatred to the large user base in social media \cite{chakravarthi-etal-2020-corpus,chakravarthi-etal-2020-sentiment}. It is plausible that memes might become an integral part of most of the people, as it is used to understand racial and gender discourse on social media platforms such as Reddit \cite{Milner2013FCJ156HT,nikhilhope,nikhiloffen}. One of the approaches to overcome this is manually monitoring and moderating user-generated content. But due to the amount of data being generated on the internet every day, it would be ideal to develop automated systems to moderate them \cite{kumar-etal-2018-benchmarking,adeepoffensive,adeephope,10.1145/3441501.3441515,10.1145/3441501.3441517}.

Consider countries with huge populations such as India, several memes are directed towards targeted communities. To address these issues of identifying if a given meme is trolling a person's sentiments, a dataset for memes that were suspected to troll a particular community. We participate in the shared task on meme classification based on the troll classification of Tamil Memes \cite{suryawanshi-etal-2020-dataset}. Tamil (ISO 639-3: tam) language is spoken in South Asia \cite{chakravarthi2020leveraging}. The earliest inscription in India dated from 580 BCE was the Tamil inscription in pottery and then the Asoka inscription in Prakrit, Greek, and Aramaic dating from 260 BCE. The earliest known inscriptions in Sanskrit are from the inscriptions of the 1st century BCE. Tamil is the official language of Tamil Nadu, India,
as well as of Singapore and Sri Lanka \cite{chakravarthi-etal-2018-improving,chakravarthi-etal-2019-wordnet}.  The task primarily consists of identifying whether a meme is a \emph{troll} or a \emph{non-troll} \cite{dravidiantrollmeme-eacl}. We use the images and captions that are provided to achieve the most efficient model to classify the memes. We use a combination of Vision Transform (ViT) \cite{dosovitskiy2021an} and mBERT \cite{pires-etal-2019-multilingual} over other pretrained models used for image classification as described in \cite{Venkatesh2020TransferLB,10.3844/jcssp.2021.44.54}.

\section{Related Work}
\label{related work}
Internet memes have been a subject of interest for both Computer Vision and Natural Language Processing researchers. The type of memes that are being used illustrates the context of discussions on social media platforms. People are using memes to express themselves, and in the making, showcase their stance on a certain social issue, be it in acknowledgment or rejection of the issue \cite{8354676,boinepelli-etal-2020-sis,Gal2016ItGB}. There exist several reasons that suggest the spread of memes. Some of the reasons include novelty, simplicity, coherence. It also includes an emotional attachment, its ability to have different meanings, depending on how a person perceives it \cite{Nave2018TalkingIP,stephens2018ryan,Chielens2002OperationalizationOM}. \citeauthor{Hu_2018} developed a multimodal sentiment analysis by developing a deep neural network that combines both visual analysis and text analysis to predict the emotional state of the user by using Tumblr posts.

\section{Data}
\label{data}
We use Troll Classification dataset of Tamil Memes \cite{suryawanshi-etal-2020-dataset}. It consists of 2,699 memes, of which most of the images have text embedded within them. We are also provided with captions for all images. The distribution is shown is Table \ref{tab1}.

\begin{table}[!h]
\begin{center}
\renewcommand{\tabcolsep}{3mm}
\begin{tabular}{|l|r|r|r|}
\hline
Class &  Train & Validation & Test\\
\hline
Troll & 1,154 & 128 & 395\\
Non-Troll & 917 & 101 & 272\\ 
\hline
total & 2,071 & 229 & 667 \\
\hline
\end{tabular}
\end{center}
\caption{Dataset Distribution}\label{tab1}
\end{table}

\section{System Description}
\label{system description}
Multimodal deep learning is a robust and efficient way of addressing the main goals of artificial intelligence by integrating and combining multiple communicative modalities to obtain crucial results which usually improves the outcome of the single models trained. As deep learning models tend to extract features on their own, the objective can easily be achieved with the help of neural networks. 

\begin{table*}[!h]
\begin{center}
\renewcommand{\tabcolsep}{3mm}
\begin{tabular}{l|r|r|r|r}
\hline
 & Precision & Recall & F1-Score & Support\\
\hline
Non-Troll & 0.96 & 0.95 & 0.96 & 101\\
Troll & 0.96 & 0.97 & 0.96 & 128\\ 
\hline
Accuracy & & & 0.96 & 229\\
Macro Avg & 0.96 & 0.96 & 0.96 & 229 \\
Weighted Avg & 0.96 & 0.96 & 0.96 & 229\\
\hline
\end{tabular}
\end{center}
\caption{Classification report of ViT to images of validation set}\label{tab2}
\end{table*}

\begin{table*}[!h]
\begin{center}
\renewcommand{\tabcolsep}{3mm}
\begin{tabular}{l|r|r|r|r}
\hline
 & Precision & Recall & F1-Score & Support\\
\hline
Non-Troll & 0.87 & 0.99 & 0.93 & 101\\
Troll & 0.99 & 0.88 & 0.93 & 128\\ 
\hline
Accuracy & & & 0.93 & 229\\
Macro Avg & 0.93 & 0.94 & 0.93 & 229 \\
Weighted Avg & 0.94 & 0.93 & 0.93 & 229\\
\hline
\end{tabular}
\end{center}
\caption{Classification report when memes are classified based on captions on validation set}\label{tab3}
\end{table*}

Given the images of Tamil Memes, along with the embedded text on the images, scrutiny of images and texts independently and then picking out relevant information for further process plays a climacteric role in our system. At the end of the training, the model has to output a single value stating the given meme is Troll or Non-Troll. The specialty of our model was to neither use the Convolutional Neural Networks (CNN) nor Recurrent Neural Networks (RNN). As the title of the paper points out, the model tries to gain more attention towards the salient portions of text and images. The proposed solution makes an effort to convey the importance of attention gain and its relation to the performance of the model. The model is put forward to compute the classification is \textbf{Vision transformer} \cite{dosovitskiy2021an} for images and \textbf{Bidirectional Encoder Representations from Transformers (BERT)} \cite{devlin-etal-2019-bert} for captions of memes. This corresponds to a \emph{transformer-transformer} architecture as shown in Fig \ref{fig1}.

\subsection{Vision Transformer (ViT)}
\label{ViT}
The architecture of the ViT is analogous to the transformer used for Natural Language Processing (NLP) tasks. NLP transformers use self-attention which is a highly cost-inefficient approach in regard to images. Admitting this, the technique applied here was Global Attention. Keeping the analogy of sentences, instead of 1D token embeddings as input, ViT receives a sequence of flattened 2D patches. If H, W is the height and width of the image and (P, P) is the resolution of each patch, \(N=HW/P^2\) is the effective sequence length for the transformer \cite{dosovitskiy2021an}. Then the patches are projected linearly and then multiplied with an embedding matrix to eventually form patched embeddings. This along with position embeddings are sent through the transformer. Similar to BERT's [CLS] token, a token is prepended along with the patched embeddings. The transformer consists of an encoder block which consists of alternating layers of multiheaded self-attention blocks to generate attention for specific regions of the images. Layer normalization and residual connections are made comparable to the original NLP transformer.

\begin{table*}[!h]
\centering
\begin{tabular}{l|r|r|r|r}
\hline
 & Precision & Recall & F1-Score & Support\\
\hline
Non-Troll & 0.60 & 0.03 & 0.06 & 272\\
Troll & 0.60 & 0.98 & 0.74 & 395\\ 
\hline 
Accuracy & & & 0.60 & 667\\
macro Avg & 0.60 & 0.51 & 0.40 & 667 \\
Weighted Avg & 0.60 & 0.60 & 0.47 & 667\\
\hline
\end{tabular}
\caption{Classification report of our system on the test set}\label{tab4}
\end{table*}
\subsection{BERT}
\label{bert}
The success of fine-tuning a pretrained model in computer-vision prompted researchers to do the same in Natural Language Processing. Therefore it was the objective of the researchers to develop a model which can be fine-tuned for NLP related works. \textbf{B}idirectional \textbf{E}ncoder \textbf{R}epresentations from \textbf{T}ransformers \textbf{(BERT)} \cite{devlin-etal-2019-bert} is a language representation model which was trained on Wikipedia corpus. The training phase had two tasks. First was Masked Language Modelling(MLM), where the sentence had random masks in them and the model has to predict the masked word. The second task Next Sentence Prediction(NSP), where the model has to predict whether the second sentence is the continuation of the first one. 

The input to the transformer is the sum of the token segmentation and positional embeddings. As the name suggests, the model is jointly conditioned on both left and right contexts to extract meaning. BERT is comparable to the transformer encoder block of \cite{NIPS2017_3f5ee243}. The NSP task matches the classification task for the objective of the model. During NSP, two sentence separated by [SEP] and [CLS] token are fed in and the output of the [CLS] token is pondered upon to determine the required class. Here, the input is only a single sentence with tokens and the model is fine-tuned as necessary. For this system,  \textit{bert-base-multilingual-cased} (L=12, H=768, A=12, Total Parameters=179M) was used. This model is pretrained on largest available Wikipedia dumps of the top 104 different languages, with the largest MLM objective, also making the model case sensitive \cite{pires-etal-2019-multilingual}.

\section{Experiments}
\label{experiments}
 All suitable models were implemented using PyTorch version 1.5.0 in a google colaboratory environment. The early stages of this model include preprocessing of images. The dataset had pictures with various resolutions and had to be made equal. The images were resized to 256 X 256 pixels. Most of the images had texts on the top and bottom of the images. Texts in the images were considered as noise for classification, which resulted in performing a center crop for all images. The border of the portions was removed and images of size 224 X 224 were produced. Finally, the images were ready as the input to the transformer by normalizing the RGB channels with mean 0.485, 0.456, 0.406, and standard deviation 0.229, 0.224, 0.225 respectively. No augmentations were made to preserve the meaning of the images. The transformer was originally trained on the ImageNet dataset and had achieved remarkable results. The trained weights are transferred to this downstream task. The base version of ViT is fine-tuned which had default hyperparameters of 16 patches, an embed dimension of 768, 12 layers, 12 attention heads, and a dropout rate of 0.1. The head of the vision transformer, which outputs 1000 classes, is now replaced by a linear layer of 128 neurons. 
 \captionsetup[sub]{labelformat=simple} 
\renewcommand{\thesubfigure}{(\alph{subfigure})}
\begin{figure}[!hbt] 
    \centering 
    \begin{subfigure}[b]{0.9\linewidth}  
        \centering 
        \includegraphics[width=\textwidth]{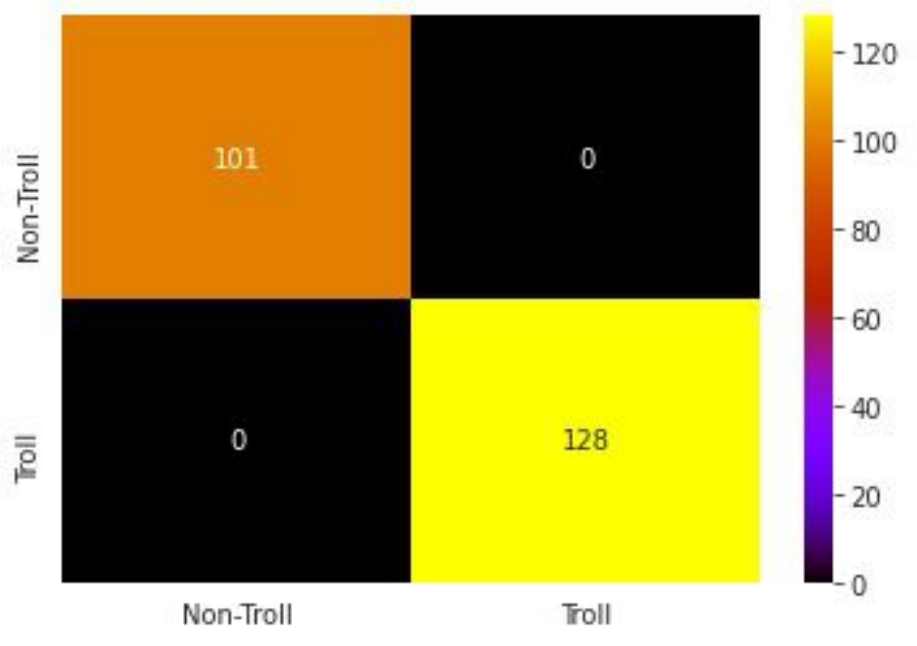} 
        \caption{Validation set} 
        \label{fig:1-1} 
    \end{subfigure} 
    \\ 
    \begin{subfigure}[b]{0.9\linewidth} 
        \centering 
        \includegraphics[width=\textwidth]{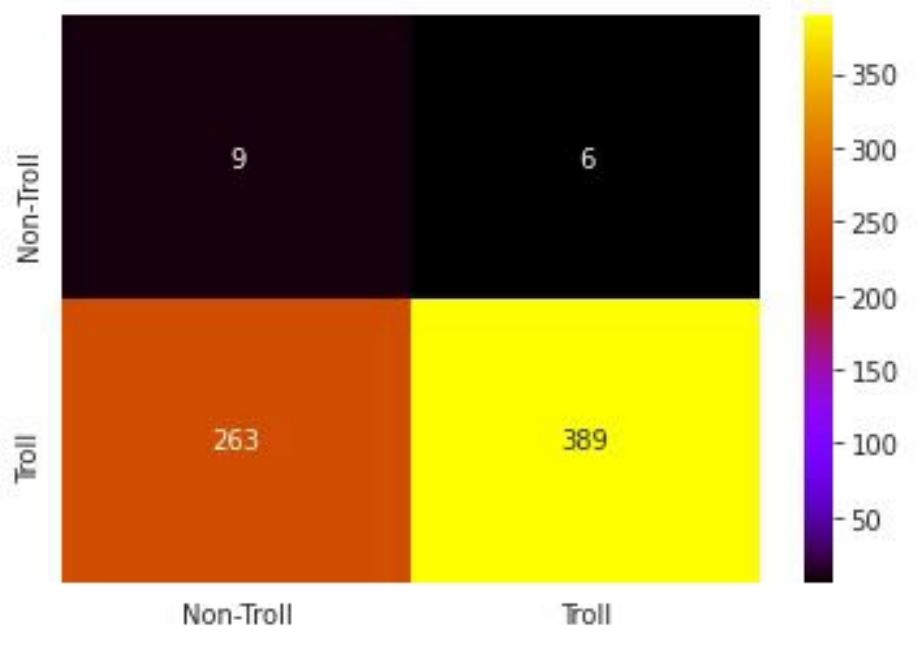} 
        \caption{ Test set} 
        \label{fig:1-2} 
    \end{subfigure} 
    \caption{Confusion Matrix} 
    \label{fig:1}  
\end{figure} 
 The texts were also preprocessed by removing stopwords, special characters, and punctuation. Texts need to be tokenized before feeding into the BERT configuration. After inserting it into the transformer, the resulting pooled output from the multilingual BERT model is also passed through a linear layer of 128 neurons.  
 
 The two layers obtained from the transformers are merged together to form a single layer with 256 neurons. This is passed through the ReLu activation function and a dropout to obtain one final neuron which determines the class as Troll or Non-Troll. A learning rate of \(2e-5\) was used with a batch size of 16. The maximum length of the captions was truncated to 128 as memes usually do not contain very long sentences. The training was done for 4 epochs and with a linear schedule with warmup. To our surprise, the model learned very rapidly and achieved well progress on the validation set which mimicked the train set. It was also observed that merging the outputs of two different domain models did not harm the training, moreover, it helped in getting better results.

\section{Results}
\label{results}
We achieve an overall F1-score of 0.96 when we use images for classification using ViT as shown in \ref{tab2}. It is to be noted that using mBERT to classify memes solely based on the captions achieves 0.93 as F1-score as shown in Table \ref{tab3}. While we achieve such good results in comparison to the baseline scores of 0.59 mentioned in the dataset paper, we feel that if both of representations of ViT and mBERT were concatenated and then fed into a linear layer, the model would learn better. We find that the model achieves a perfect 1.00 weighted F1-score on the validation set. We believe that preprocessing of the images was a major factor for achieving a great F1-score on validation set. This argument is supported by our system's poor performance on the test set, as the test set was not coherent with the training data in terms of the positioning of texts on the images as shown in Table \ref{tab4}. The confusion matrix on validation and test set are as shown in Figures \ref{fig:1-1} and \ref{fig:1-2} respectively.

\section{Conclusions}
\label{conclusion}
The proposed solution performs at greater heights on the validation and set in the training phase. The validation set mimics the train set as the memes are split looking at the distribution of the classes. The dataset is very small and augmenting it will not help for the optimal results. The algorithm over-fits the train set undoubtedly. The reason behind the poor performance is due to the change in the distribution. The memes in the test set had multiple images which were difficult for the ViT to capture features. The model scored a F1 score of 0.46 on the test set and 1.0 on the validation set. Vast difference can be observed due to high bias. Here, in this paper, we have tried to come up with this innovation of transformer-transformer architecture which can achieve extreme results. In the future, we will be performing a wonderful task of having more transformers in parallel computation and syncing them makes an immense difference in this era of deep learning.





 .




\bibliography{anthology,eacl2021}
\bibliographystyle{acl_natbib}

\end{document}